\documentclass{article} % For LaTeX2e
\usepackage{amssymb}
\usepackage{pifont}
\usepackage{textcomp}

\usepackage{iclr2024_conference, times}
\usepackage{amsmath,amsfonts,bm}

\def\eqref#1{equation~\ref{#1}}
\def\1{\bm{1}}

\DeclareMathAlphabet{\mathsfit}{\encodingdefault}{\sfdefault}{m}{sl}
\SetMathAlphabet{\mathsfit}{bold}{\encodingdefault}{\sfdefault}{bx}{n}

\usepackage{graphicx}
\usepackage{booktabs} % for professional tables
\usepackage{dblfloatfix}    % To enable figures at the bottom of page
\usepackage{float}
\usepackage{hyperref} % hyperref makes hyperlinks in the resulting PDF.
\usepackage{url}
\usepackage{subcaption} 
\usepackage[normalem]{ulem}

\title{Generalizability Under Sensor Failure: \\Tokenization + Transformers Enable More Robust Latent Spaces}

\author{Geeling Chau\thanks{ Equal Contribution }, Yujin An\footnotemark[1], Ahamed Raffey Iqbal\footnotemark[1], \textbf{Soon-Jo Chung, Yisong Yue, Sabera Talukder}  \\
California Institute of Technology, Pasadena, CA \\
\texttt{\{gchau,yan2,raffey,sjchung,yyue,sabera\}@caltech.edu}
}

\iclrfinalcopy % Uncomment for camera-ready version, but NOT for submission.
\begin{document}

\noindent\maketitle

\setlength\parindent{24pt}

% SABERA COMMENTS:
% Use cite, citep (citep is cite parenthetical) —> and be particular when you use one versus the other 
% Your abstract should not have any citations if it can be avoided --> there is one piece where I dont think it can be avoided
% Your abstract should not have your contributions explicitly - it should instead have the nugget(s) of wisdom you want the reader to gain with potentially some supporting evidence for it

\begin{abstract}
A major goal in neuroscience is to discover neural data representations that generalize. This goal is challenged by variability along recording sessions (e.g. environment), subjects (e.g. varying neural structures), and sensors (e.g. sensor noise), among others. Recent work has begun to address generalization across sessions and subjects, but few study robustness to sensor failure which is highly prevalent in neuroscience experiments. In order to address these generalizability dimensions we first collect our own electroencephalography dataset with numerous sessions, subjects, and sensors, then study two time series models: EEGNet \citep{lawhern2018eegnet} and TOTEM \citep{talukder2024totem}. EEGNet is a widely used convolutional neural network, while TOTEM is a discrete time series tokenizer and transformer model. We find that TOTEM outperforms or matches EEGNet across all generalizability cases. Finally through analysis of TOTEM's latent codebook we observe that tokenization enables generalization.

\end{abstract}

\section{Introduction}
Neuroscience experiments vary across numerous dimensions including sessions, subjects and senors \citep{gonschorek2021removing, saha2020intra, parvizi2018promises}. Given this inherent variability, models with strong generalization properties are desirable. Model generalizability refers to a model's zero shot capabilities, or the model's ability to operate on test datasets unseen at training time. Prior work studies generalizability along the datasets' session and subject dimensions \citep{peterson2021generalized, talukder2022deep, krumpe2017non}, but few study model generalizability under sensor variability. Sensors vary primarily because of sensor failure and sensor count differences across recording sessions. Common practice is to reduce the train and test sets to the intersection of clean and available sensors then train models. However, this throws away difficult-to-obtain neural data. It is therefore valuable to build models that can train on all available data and infer under any type of sensor variability.

\begin{figure}[ht]
    \centering
    \includegraphics[width=0.75\linewidth]{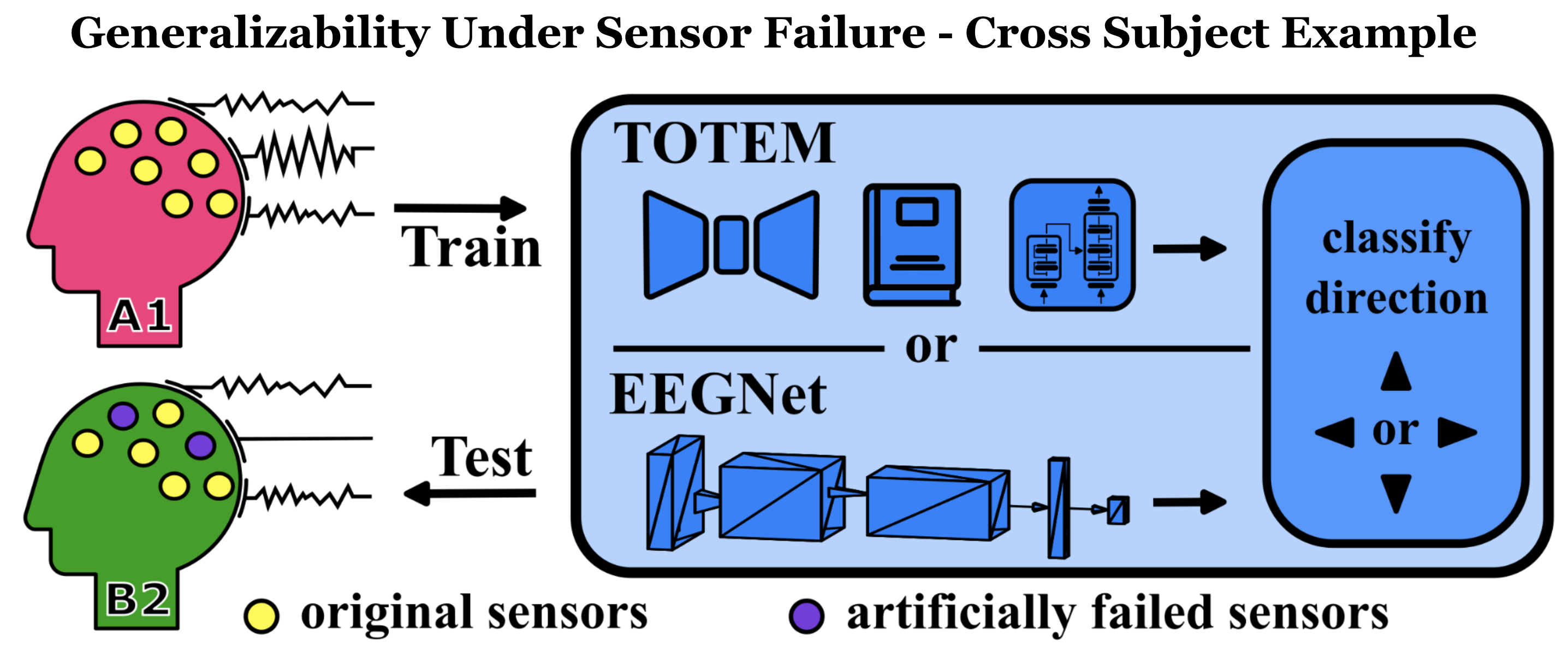}
    \caption{\textbf{Overview}. TOTEM and EEGNet train on data from subject A1, which has no failed sensors. Both TOTEM and EEGNet are then tested on subject B2 with artificially failed sensors. This is an example of cross subject generalizability under sensor failure.}
    \label{fig:intro}
\end{figure}

To probe generalizability across sessions, subjects, and sensors, we systematically study two time series models: EEGNet \citep{lawhern2018eegnet} and TOTEM \citep{talukder2024totem}. EEGNet is a popular convolutional neural network upon which many other models are built or compared against \citep{peterson2021generalized, xu2021enhancing}. EEGNet intakes a Sensor x Time array and applies convolutional kernels along the temporal and spatial dimensions, a common approach inspired by Filter-Bank Common Spatial Patterns (FBCSP) \cite{ang2012filter}. TOTEM first learns a sensor-agnostic set of discrete tokens via a self-supervised vector quantized variational autoencoder, then uses these pretrained tokens as the input to a transformer classifier. TOTEM exhibits strong generalization on numerous time series datasets \citep{talukder2024totem}. To study these models' generalizability across experimental conditions we create a taxonomy of generalization cases that encompass session, subject, and sensor variability. We collect a rich electroencephalography dataset that permits testing of these generalization cases.
See Figure~\ref{fig:intro} for a visualization of cross subject generalizability under sensor failure. Finally we explore TOTEM's latent discrete codebooks, as TOTEM demonstrates the best performance across experiments in our generalization taxonomy.

\section{Metholodology}
\label{gen_inst}

\subsection{Experimental design}
\begin{figure}[h!]
    \centering
    \begin{subfigure}[t]{0.3\textwidth}
        \includegraphics[width=\hsize]{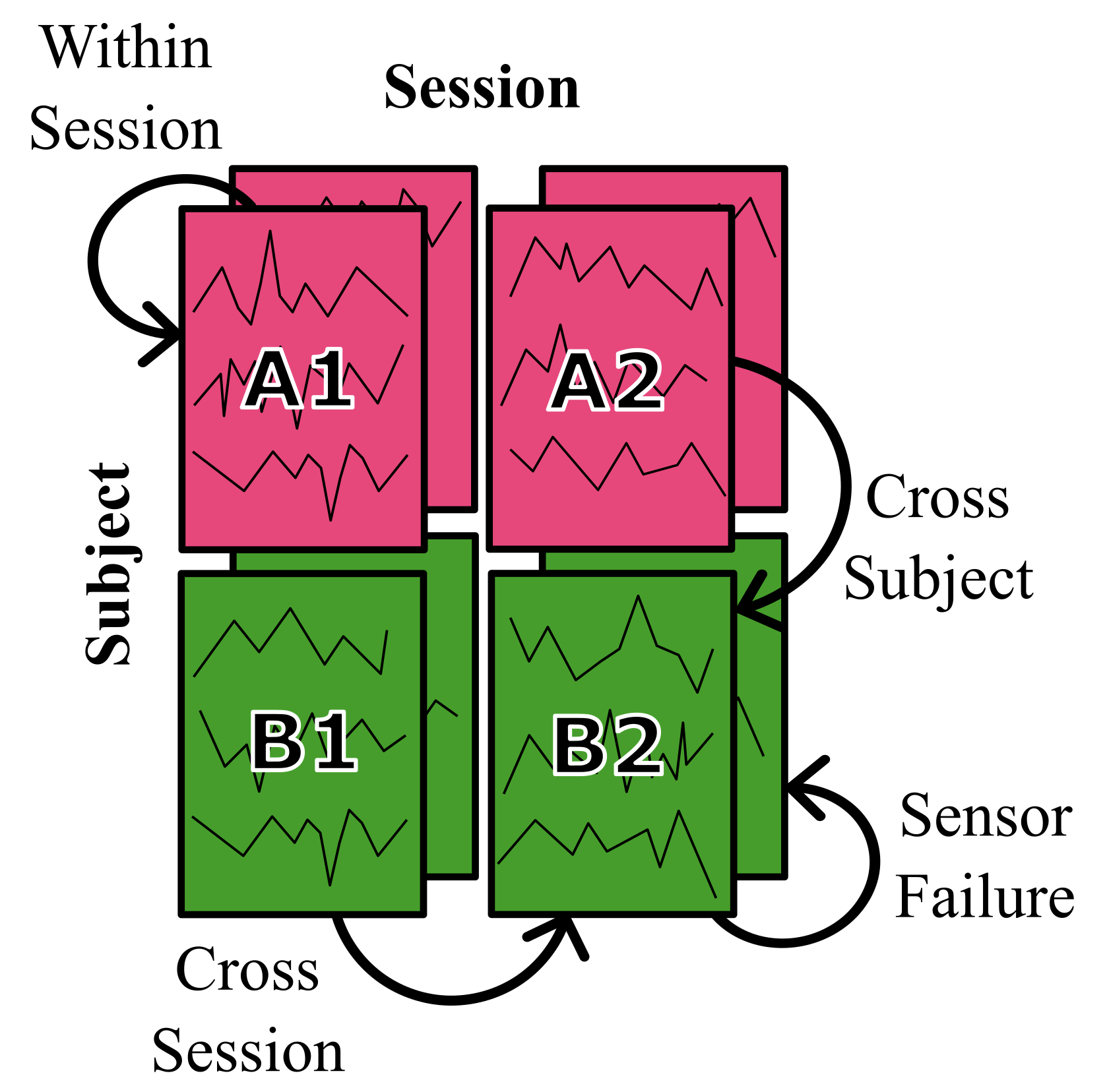}
        \caption{}
        \label{fig:dataset_1}
    \end{subfigure}
    \begin{subfigure}[t]{0.45\textwidth}
        \includegraphics[width=\hsize]{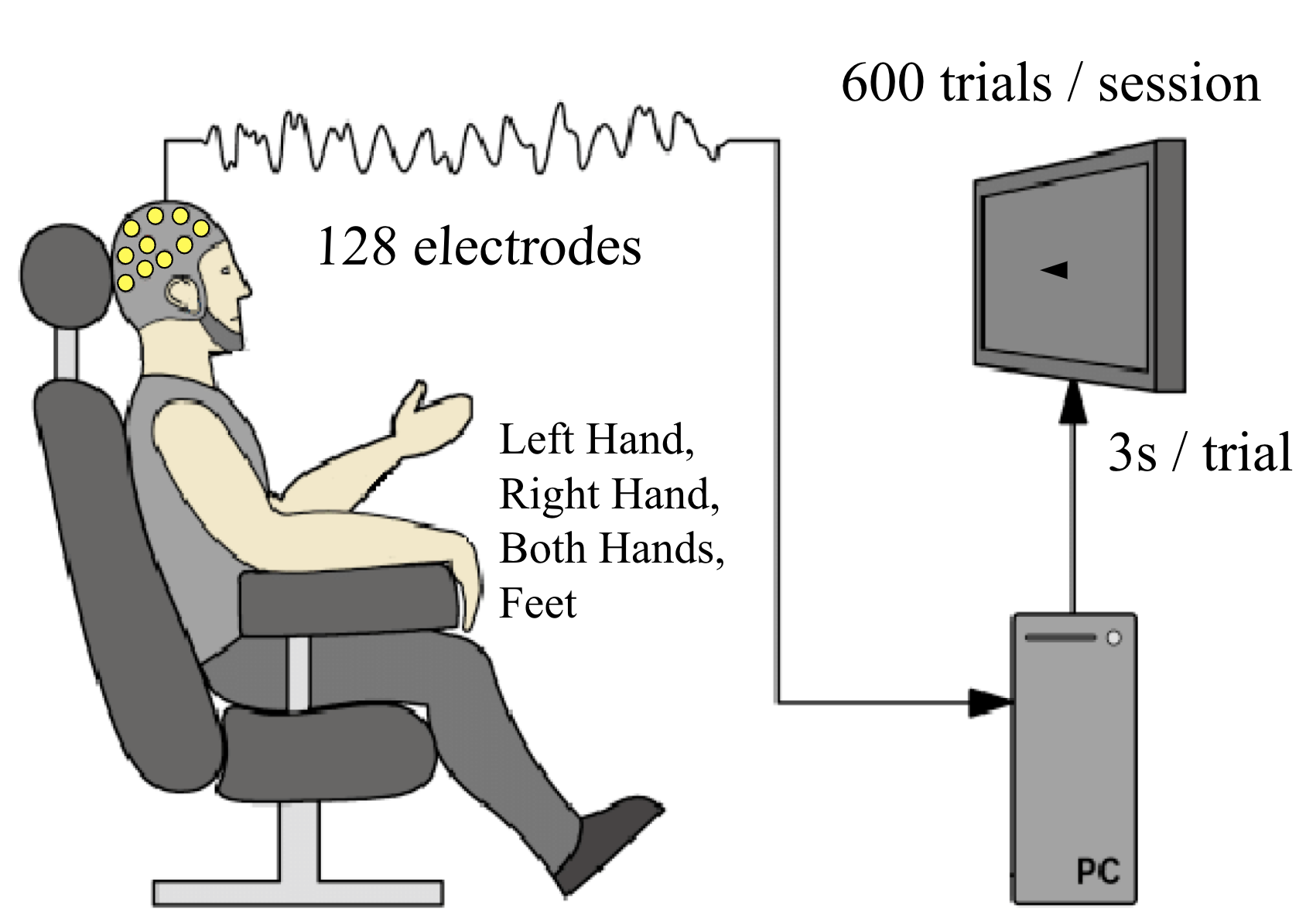}
        \caption{}
        \label{fig:dataset_2}
    \end{subfigure}
    \caption{(a) Visualization of baseline (within session) and generalizability cases (cross subject, cross session, sensor failure) illustrated. (b) Experimental setup. We test on 2 human subjects each with 128 electrodes generating 600 trials per session. Adapted from \cite{garcia2023kcs}}
    \label{fig:dataset}
\end{figure}

We first define a taxonomy of a baseline and generalization cases, the cases are as follows: \underline{\textbf{Baseline Case:}} \emph{Within Session} - train and test on the same session. \underline{\textbf{Generalization Cases:}} (1) \emph{Cross Session} - train on one session, and test on a separate session from the same subject. (2) \emph{Cross Subject} - train on one subject, and test on a separate subject. (3) \emph{Sensor Failure} - randomly fail X\% of sensors, $X \in \{0,10,20, ...,100\}$. Notably the within session, cross session, and cross subject cases can be combined with the sensor failure condition, e.g. one can study cross subject generalizability under sensor failure (Figure~\ref{fig:intro}). For a visualization of the baseline and generalizability cases see Figure \ref{fig:dataset}(a).

We then design and collect an electroencepholography (EEG) dataset featuring high sensor count (128 electrodes) and high trial count (600 trials/session) across two human subjects. In total this leads to four sub-datasets across subjects A,B and sessions 1,2: A1, A2, B1, B2. Our 128 sensor setup introduces relatively large datasets as previous publicly available EEG datasets rarely exceed 64 sensors \citep{stieger2021continuous, kaya2018large, tangermann2012review}. Additionally, much prior work with more than 64 channels collects \textless200 trials per session \citep{gwon2023review, xu2020cross} we collect 150 trials per direction (600 trials per session). Designing our experiments to generate relatively large datasets enables our study of generalization across numerous experimental conditions.

For each session a subject sat in front of a monitor and fixated on a center point that randomly changed to $\blacktriangleleft$, $\blacktriangleright$, $\blacktriangle$, and $\blacktriangledown$; $\blacktriangleleft$ is left hand movement (HM), $\blacktriangleright$ is right HM, $\blacktriangle$ is both HM, and $\blacktriangledown$ is both feet movement. In each session there were 150 trials for each direction, lasting 3 seconds each. After data collection we performed minimal data processing by (1) downsampling from 4096Hz to 2048Hz, (2) high-pass filtering at 0.5Hz, (3) average referencing, and (4) standardizing across the entire recording.

We chose to use EEG due to its high variance along these generalization cases, allowing us to study generalizability in one of the most difficult data modalities. However, challenges regarding sensor variation are highly prominent in larger signal-to-noise ratio recording modalities such as electrocorticography or multi-unit probes (Neuropixels, Utah arrays, DBS probes). In the future we aim to replicate our EEG experiments with other neural data recording modalities, see Section~\ref{conc} for further discussion.

\subsection{Models}

\begin{figure}[h!]
    \centering
    \includegraphics[width=0.6\linewidth]{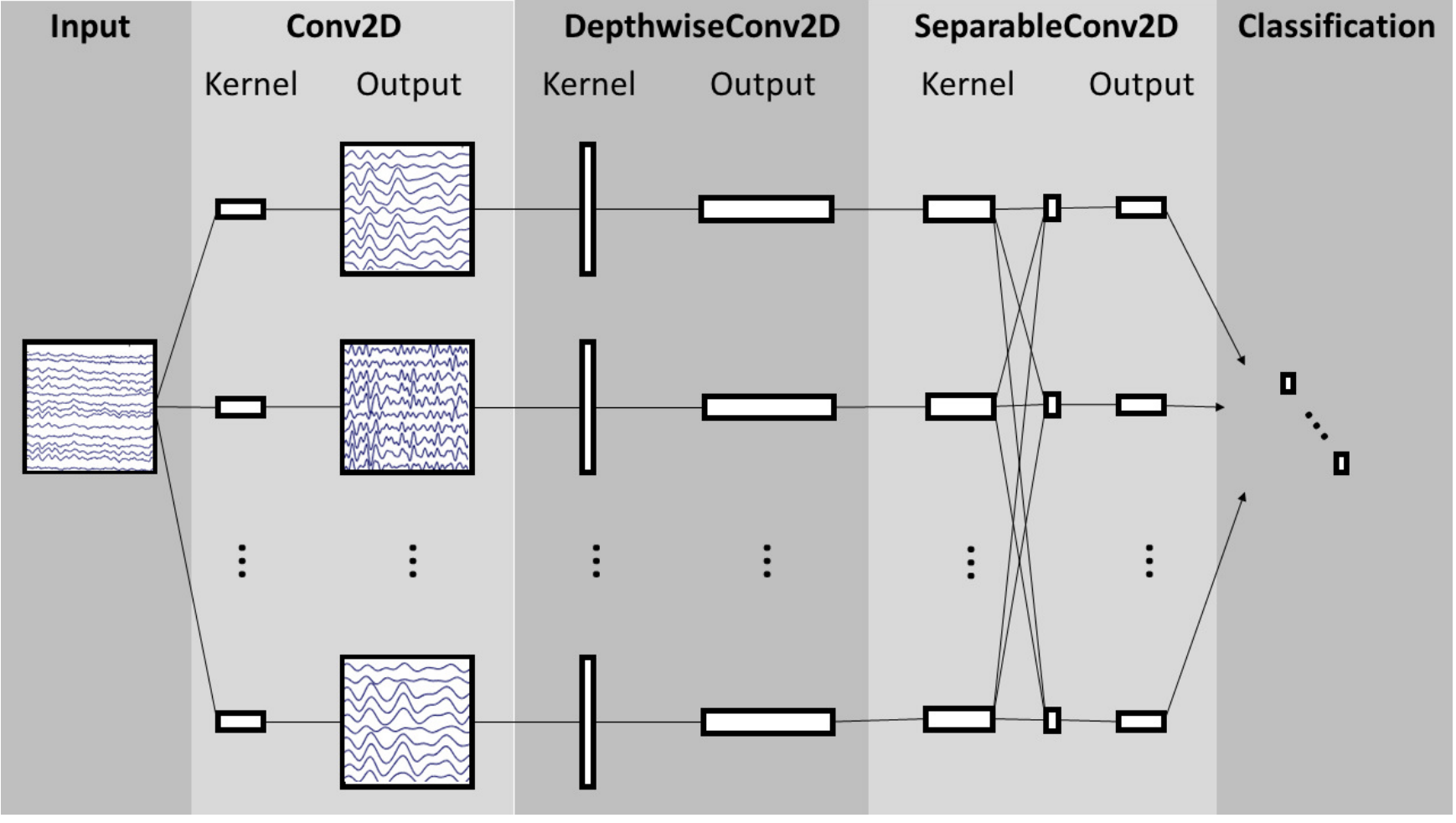}
    \caption{EEGNet architecture. Adapted from \cite{lawhern2018eegnet}}
    \label{fig:eegnet}
\end{figure}

EEGNet from \cite{lawhern2018eegnet} is a popularly used convolutional neural network (CNN) for EEG decoding. It consists of temporal and spatial CNNs that learn kernels swept across these dimensions. Its design mimics a performant EEG processing pipeline, FBCSP (\cite{ang2012filter}), learning temporal kernels that extract frequency features which then get weighted by learned spatial filters (Figure \ref{fig:eegnet}). EEGNet takes a fixed [Sensors x Time] matrix as input. 

\begin{figure}[h!]
    \centering
    \includegraphics[width=0.6\linewidth]{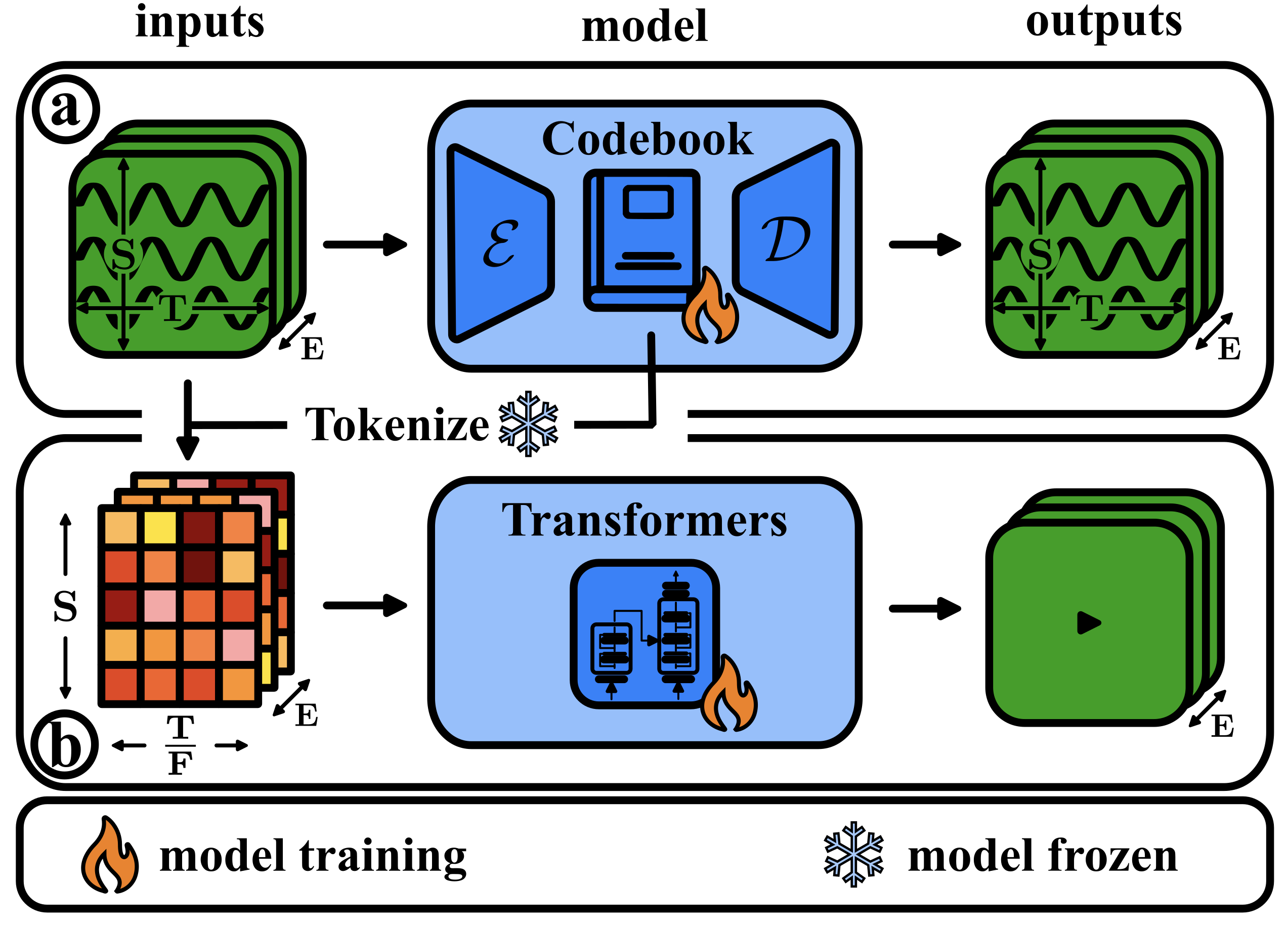}
    \caption{TOTEM architecture and training. (a) Learn latent codebok via self-supervision. (b) Train transformers on tokenized data created from frozen codebook. Adapted from \cite{talukder2024totem}}
    \label{fig:totem}
\end{figure}

TOTEM from \cite{talukder2024totem} consists of a sensor-agnostic tokenizer which uses a learned latent codebook, followed by a time-length flexible and sensor-count flexible transformer classifier (Figure \ref{fig:totem}). TOTEM first learns the latent codebook via self-supervision on the time series signals (Figure \ref{fig:totem}(a)). Its goal is to learn a set of small time tokens that help reconstruct the time series signals with lowest MSE. This can be seen as learning a discrete temporal ``vocabulary" that can be used to express the data modality's distribution of temporal activity. This tokeniation technique is inspired by the VQ-VAE (\cite{van2017neural}) which has been studied to be beneficial in helping (1) reduce noise in sequential representations and (2) allow transformers to operate on a meaningful and defined vocabulary set \citep{talukder2024totem}. After the codebook training has converged, it is then frozen and used to translate the time series inputs into tokenized inputs for the downstream temporal and spatial transformers (Figure \ref{fig:totem}(b)). These transformers are able to operate on different sequence lengths (\cite{vaswani2017attention}), providing benefits for modeling trials with different lengths and sensor availabilities. We here only study the case where the input sizes are not changed in order to be able to compare with EEGNet.

\subsection{Training}

The 600 trials in each session are broken down into fixed training (80\%), validation (10\%), and test (10\%) sets, within each of which are then broken down into 250ms long individual mini-trials. We train and test on these mini-trials. In the baseline within session case, we train, val, and test on the same session. In generalization cross session and subject cases, we train and validate on the first session, then test on the second session. In sensor failure cases, we train and validate on the original sensors and test on a set with artificially failed sensors.

Sensor Failure: We simulate failure by randomly zeroing-out X\% of test set sensors, where X$\in\{0,10,20, ...,100\}$ (11 cases). We cannot remove sensors as EEGNet requires fixed input size, so we zero-out sensors for both models. For fair comparison, we ensure both models have the same failed sensors for each random seed. We study the effect on classification performance across within session and generalizability cases.

Hyperparameter selection: For each model, we selected a set of hyperparameters that allowed the models to converge and perform optimally on within session performance. We fixed these hyperparameters and ran all our modeling experiments with the same parameters across datasets.

\section{Results \& Discussion}
\label{headings}

We report decoding accuracy across all generalization cases (Figure \ref{fig:decoding_accuracy}) and analyze TOTEM’s latent spaces to understand how its representations generalize (Figure \ref{fig:codebook_analysis}).

\subsection{Decoding Accuracy Comparison}
To evaluate the robustness of the latent spaces learned by each of the models, we tested their performance on the baseline case (within session) and zero-shot to generalization cases (cross session, cross subject, and across sensor failure levels). 

\begin{figure}[h!]
    \centering
    \includegraphics[width=\linewidth]{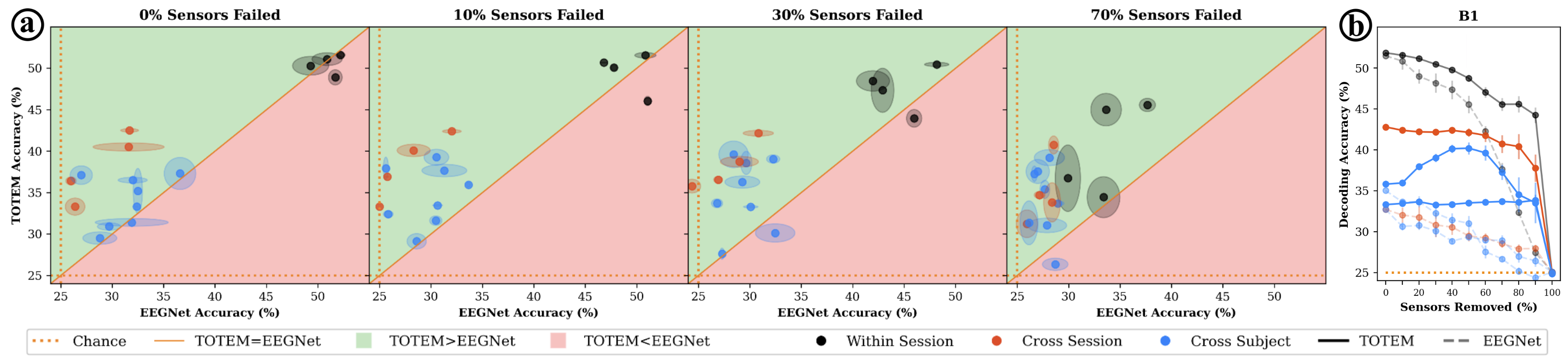}
    \caption{Classifier performance under all generalizability cases. (a) TOTEM vs EEGNet accuracy for all within session (black), cross session (red), cross subject (blue) cases across several amounts of sensor failure (0\%, 10\%, 30\%, 70\%). Ovals in 0\% failure case represent standard error of mean (SEM) across 5 model random seeds. Ovals in 10\%, 30\%, 70\% sensor failure plots represent SEM across 3 sensor failure random seeds. (b) Decoding accuracy of TOTEM (solid line) and EEGNet (dashed line) when trained on B1, and tested against within session (black), cross session (red), and cross subject (blue) cases across 0-100\% sensor failure percentages. Additional performances can be found in Appendix \ref{appendix:sensor_failure_detailed}.}
    \label{fig:decoding_accuracy}
\end{figure}

When analyzing the 0\% sensor failure generalizability cases, we see TOTEM outperforms EEGNet on cross session and subject cases, and match EEGNet on within session performance. On a case by case basis, we see that TOTEM beats EEGNet on all cross session modeling experiments, and beats in 4/8 cross subject modeling experiments. The within session experiments all cluster around the equal performance line, which means that TOTEM is able to do just as well as EEGNet when evaluated at the baseline case. The difference in generalizability performance between the two models while within session decoding is fairly consistent shows that it is important to evaluate models along generalization dimensions when selecting for models.

When analyzing the 10\%, 30\%, and 70\% generalizability cases, we see that for nearly all modeling experiments, the same generalizbility results we saw hold, with TOTEM's performance becoming relatively better than EEGNet's as more sensors are failed. This can be observed by the dots shifting leftward and more into the green regions as more sensors are failed (Figure \ref{fig:decoding_accuracy}). The within session decoding accuracies also favor TOTEM as more sensors are failed, which is beneficial for modeling within session datasets where sensor failure may appear randomly within a session.

When analyzing the trends of each model against a finer resolution of sensor failure percentages, we see that TOTEM maintains a higher decoding accuracy for longer, while EEGNet has a more linear declining performance with additional sensors removed (Figure \ref{fig:decoding_accuracy} (b)). This is observed even when both models start with the same decoding accuracy, so it is not due to TOTEM simply generalizing better in cross session and cross subject cases. Additional results trained on each of our 4 sessions can be found in Appendix \ref{appendix:sensor_failure_detailed}.

These results suggest that (1) TOTEM’s tokenization + transformers approach create more generalizable representations when compared to EEGNet’s CNN kernels, and (2) studying generalization highlights model performance differences as the within session conclusions differ from the generalizability conclusions.

\subsection{Latent Codebook Analysis}

\begin{figure}[h!]
    \centering
    \includegraphics[width=\linewidth]{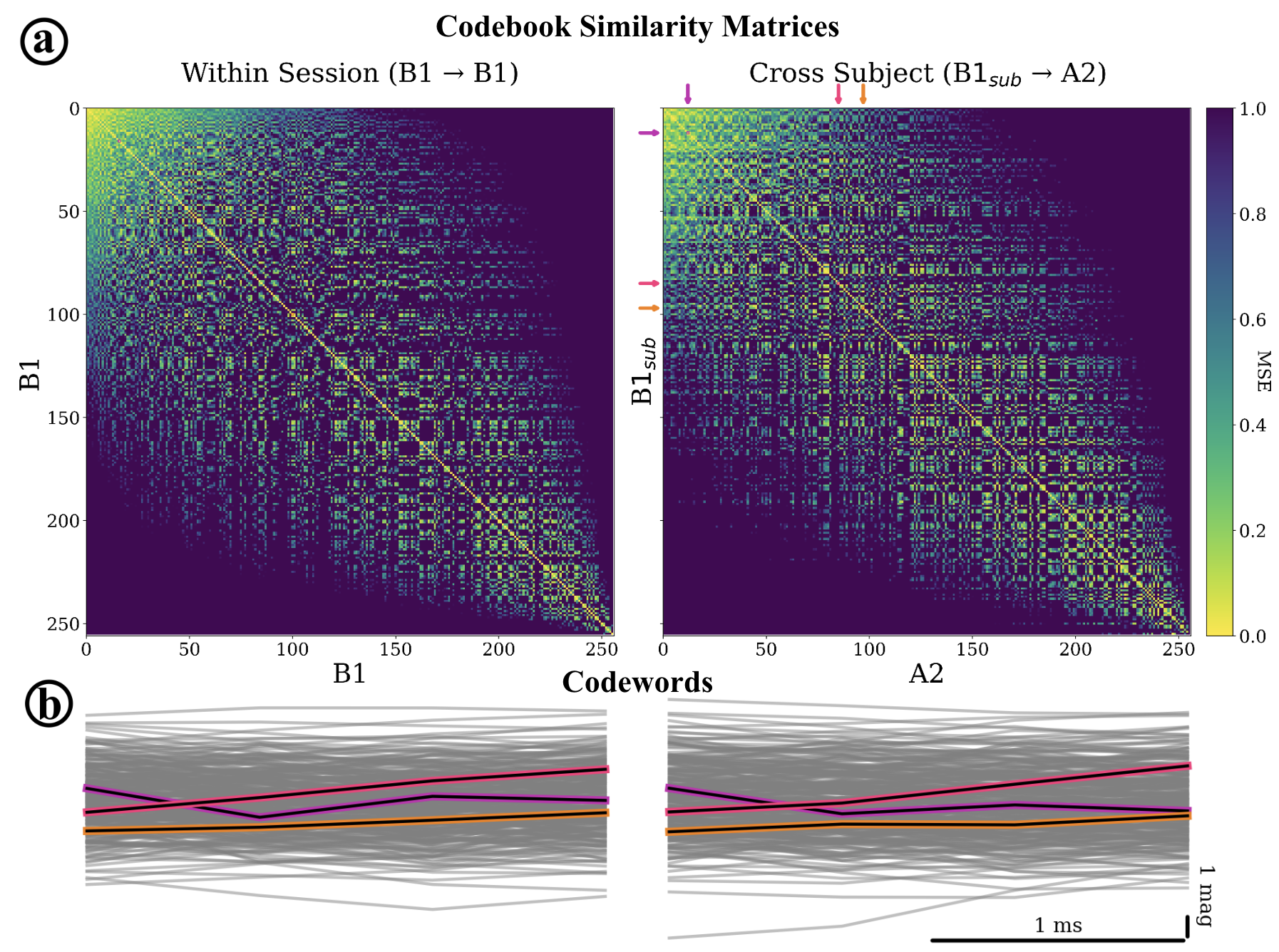}
    \caption{Codebook similarity matrices and visualization of codewords within codebooks. (a) Matrices of MSE between matching codewords in the two labeled codebooks. Codewords are ordered roughly by within session mean MSE. Colored arrows along axis denote visualized codewords highlighted in (b). (b) All codewords in each codebook. Highlighted are three examples of matched codewords between codebooks. MSE values between codewords are: purple=0.065, pink=0.018, orange=0.013.}
    \label{fig:codebook_analysis}
\end{figure}

To better understand TOTEM's generalization capabilities, we investigate its learned latent codebook across sessions, subjects, and sensor availability. If the codebooks learned are generalizable, we would expect that the codewords have corresponding matches to the codewords in the other codebooks, allowing a new recording session to be reconstructed and represented with high fidelity. Indeed, when we plot the MSE between codewords across different codebooks (Figure \ref{fig:codebook_analysis}(a) Cross Subject), we find that the  codebook similarity matrices are qualitatively comparable to within codebook (Figure \ref{fig:codebook_analysis}(a), Within Session), especially when we match each codeword from the new codebook (A2) with its minimum MSE codeword in the original codebook (B1). The matched codewords also look highly similar across codebooks (Figure \ref{fig:codebook_analysis}(b)) and are quantitatively low in MSE. Across the whole codebook, the average MSE of matched codewords is near zero and a majority of codewords are used when matching (Table \ref{table:codebook_metrics}), suggesting high overlap between codebooks learned across generalizability cases. All generalization cases studied here learn similar codebooks according to these qualitative and quantitative metrics (additional visualizations in Appendix \ref{appendix:codebook_generalizability}). This shows that TOTEM's learned tokenization space may be able to learn a generalized codebook for our time series data modality of interest, potentially allowing for mass training of the downstream transformer encoders using the same tokenized latent space.

\begin{table}[h!]
\begin{center}
\begin{tabular}{  c | c | c   }
  & Average MSE & Num. Subselected Codewords  \\ 
 \hline\hline
 Baseline Case & & \\
 \hline
 Within Session: B1 $\rightarrow$ B1 & 0.0 & 256  \\
 \hline
  Generalization Cases & & \\
 \hline
 Cross Session: B$1_{sub}$ $\rightarrow$ B2 & 0.06 & 163 \\
 Cross Subject: B$1_{sub}$ $\rightarrow$ A2 & 0.05 & 188 \\
 10\% sensors: B$1_{sub}$ $\rightarrow$ B1 10\% & 0.04 & 194 \\
\end{tabular}
\caption{Average MSE of matched codewords in baseline case and generalization cases. Number of Subselected Codewords denote the number of codewords from the original codebook (B1) that were matched to the generalization codebook. }
\label{table:codebook_metrics}
\end{center}
\end{table}

\section{Conclusion + Limitations + Future Work}\label{conc}

We find that tokenization + transformers are a promising approach to modeling time series neural data which have high variability in datasets and recordings. Specifically, we demonstrate that compared with one of the most performant and popular CNN models, tokenization+transformers outperform in numerous generalization cases. These models are also ripe for interpretability analysis which can uncover new findings about time series neural data. Our study further shows how important it is to consider generalizability cases when selecting models. 

 Currently we only model sensor failure as zeroing out data. This masking technique is widely used machine learning, but does not cover all sensor failure cases. Creating more sensor failure modes is a meaningful direction for future work. Extending our interpretability analysis to the downstream transformer is another valuable future direction. We would also like to adapt the framework to work with more types of neural time series data such as sparsely sampled neural time series data such as LFP from neuropixels, Utah Arrays, and stereoencepholography (sEEG) recordings, and more varied layout modalities such as electrocortiography (ECoG). This work could enable automatic noisy sensor detection, interesting interpretations of temporal and spatial dimensions, and building of foundation models for neural time series.

\clearpage

\bibliography{sources}

\begin{thebibliography}{18}
\providecommand{\natexlab}[1]{#1}
\providecommand{\url}[1]{\texttt{#1}}
\expandafter\ifx\csname urlstyle\endcsname\relax
  \providecommand{\doi}[1]{doi: #1}\else
  \providecommand{\doi}{doi: \begingroup \urlstyle{rm}\Url}\fi

\bibitem[Ang et~al.(2012)Ang, Chin, Wang, Guan, and Zhang]{ang2012filter}
Kai~Keng Ang, Zheng~Yang Chin, Chuanchu Wang, Cuntai Guan, and Haihong Zhang.
\newblock Filter bank common spatial pattern algorithm on bci competition iv datasets 2a and 2b.
\newblock \emph{Frontiers in neuroscience}, 6:\penalty0 21002, 2012.

\bibitem[Garc{\'\i}a-Murillo et~al.(2023)Garc{\'\i}a-Murillo, {\'A}lvarez-Meza, and Castellanos-Dominguez]{garcia2023kcs}
Daniel~Guillermo Garc{\'\i}a-Murillo, Andr{\'e}s~Marino {\'A}lvarez-Meza, and Cesar~German Castellanos-Dominguez.
\newblock Kcs-fcnet: Kernel cross-spectral functional connectivity network for eeg-based motor imagery classification.
\newblock \emph{Diagnostics}, 13\penalty0 (6):\penalty0 1122, 2023.

\bibitem[Gonschorek et~al.(2021)Gonschorek, H{\"o}fling, Szatko, Franke, Schubert, Dunn, Berens, Klindt, and Euler]{gonschorek2021removing}
Dominic Gonschorek, Larissa H{\"o}fling, Klaudia~P Szatko, Katrin Franke, Timm Schubert, Benjamin Dunn, Philipp Berens, David Klindt, and Thomas Euler.
\newblock Removing inter-experimental variability from functional data in systems neuroscience.
\newblock \emph{Advances in Neural Information Processing Systems}, 34:\penalty0 3706--3719, 2021.

\bibitem[Gwon et~al.(2023)Gwon, Won, Song, Nam, Jun, and Ahn]{gwon2023review}
Daeun Gwon, Kyungho Won, Minseok Song, Chang~S Nam, Sung~Chan Jun, and Minkyu Ahn.
\newblock Review of public motor imagery and execution datasets in brain-computer interfaces.
\newblock \emph{Frontiers in human neuroscience}, 17:\penalty0 1134869, 2023.

\bibitem[Kaya et~al.(2018)Kaya, Binli, Ozbay, Yanar, and Mishchenko]{kaya2018large}
Murat Kaya, Mustafa~Kemal Binli, Erkan Ozbay, Hilmi Yanar, and Yuriy Mishchenko.
\newblock A large electroencephalographic motor imagery dataset for electroencephalographic brain computer interfaces.
\newblock \emph{Scientific data}, 5\penalty0 (1):\penalty0 1--16, 2018.

\bibitem[Krumpe et~al.(2017)Krumpe, Baumgaertner, Rosenstiel, and Sp{\"u}ler]{krumpe2017non}
Tanja Krumpe, Katrin Baumgaertner, Wolfgang Rosenstiel, and Martin Sp{\"u}ler.
\newblock Non-stationarity and inter-subject variability of eeg characteristics in the context of bci development.
\newblock In \emph{7th Graz Brain-Computer Interface Conference 2017, Graz, Austria}. Verlag der TU Graz, 2017.

\bibitem[Lawhern et~al.(2018)Lawhern, Solon, Waytowich, Gordon, Hung, and Lance]{lawhern2018eegnet}
Vernon~J Lawhern, Amelia~J Solon, Nicholas~R Waytowich, Stephen~M Gordon, Chou~P Hung, and Brent~J Lance.
\newblock Eegnet: a compact convolutional neural network for eeg-based brain--computer interfaces.
\newblock \emph{Journal of neural engineering}, 15\penalty0 (5):\penalty0 056013, 2018.

\bibitem[Parvizi \& Kastner(2018)Parvizi and Kastner]{parvizi2018promises}
Josef Parvizi and Sabine Kastner.
\newblock Promises and limitations of human intracranial electroencephalography.
\newblock \emph{Nature neuroscience}, 21\penalty0 (4):\penalty0 474--483, 2018.

\bibitem[Peterson et~al.(2021)Peterson, Steine-Hanson, Davis, Rao, and Brunton]{peterson2021generalized}
Steven~M Peterson, Zoe Steine-Hanson, Nathan Davis, Rajesh~PN Rao, and Bingni~W Brunton.
\newblock Generalized neural decoders for transfer learning across participants and recording modalities.
\newblock \emph{Journal of Neural Engineering}, 18\penalty0 (2):\penalty0 026014, 2021.

\bibitem[Saha \& Baumert(2020)Saha and Baumert]{saha2020intra}
Simanto Saha and Mathias Baumert.
\newblock Intra-and inter-subject variability in eeg-based sensorimotor brain computer interface: a review.
\newblock \emph{Frontiers in computational neuroscience}, 13:\penalty0 87, 2020.

\bibitem[Stieger et~al.(2021)Stieger, Engel, and He]{stieger2021continuous}
James~R Stieger, Stephen~A Engel, and Bin He.
\newblock Continuous sensorimotor rhythm based brain computer interface learning in a large population.
\newblock \emph{Scientific Data}, 8\penalty0 (1):\penalty0 98, 2021.

\bibitem[Talukder et~al.(2022)Talukder, Sun, Leonard, Brunton, and Yue]{talukder2022deep}
Sabera Talukder, Jennifer~J Sun, Matthew Leonard, Bingni~W Brunton, and Yisong Yue.
\newblock Deep neural imputation: A framework for recovering incomplete brain recordings.
\newblock \emph{arXiv preprint arXiv:2206.08094}, 2022.

\bibitem[Talukder et~al.(2024)Talukder, Yue, and Gkioxari]{talukder2024totem}
Sabera Talukder, Yisong Yue, and Georgia Gkioxari.
\newblock Totem: Tokenized time series embeddings for general time series analysis.
\newblock \emph{arXiv preprint arXiv:2402.16412}, 2024.

\bibitem[Tangermann et~al.(2012)Tangermann, M{\"u}ller, Aertsen, Birbaumer, Braun, Brunner, Leeb, Mehring, Mueller-Putz, Nolte, et~al.]{tangermann2012review}
Michael Tangermann, Klaus-Robert M{\"u}ller, Ad~Aertsen, Niels Birbaumer, Christoph Braun, Clemens Brunner, Robert Leeb, Carsten Mehring, Gernot Mueller-Putz, Guido Nolte, et~al.
\newblock Review of the bci competition iv.
\newblock \emph{Frontiers in neuroscience}, 6:\penalty0 21084, 2012.

\bibitem[Van Den~Oord et~al.(2017)Van Den~Oord, Vinyals, et~al.]{van2017neural}
Aaron Van Den~Oord, Oriol Vinyals, et~al.
\newblock Neural discrete representation learning.
\newblock \emph{Advances in neural information processing systems}, 30, 2017.

\bibitem[Vaswani et~al.(2017)Vaswani, Shazeer, Parmar, Uszkoreit, Jones, Gomez, Kaiser, and Polosukhin]{vaswani2017attention}
Ashish Vaswani, Noam Shazeer, Niki Parmar, Jakob Uszkoreit, Llion Jones, Aidan~N Gomez, {\L}ukasz Kaiser, and Illia Polosukhin.
\newblock Attention is all you need.
\newblock \emph{Advances in neural information processing systems}, 30, 2017.

\bibitem[Xu et~al.(2020)Xu, Xu, Ke, An, Liu, and Ming]{xu2020cross}
Lichao Xu, Minpeng Xu, Yufeng Ke, Xingwei An, Shuang Liu, and Dong Ming.
\newblock Cross-dataset variability problem in eeg decoding with deep learning.
\newblock \emph{Frontiers in human neuroscience}, 14:\penalty0 103, 2020.

\bibitem[Xu et~al.(2021)Xu, Xu, Ma, Wang, Jung, and Ming]{xu2021enhancing}
Lichao Xu, Minpeng Xu, Zhen Ma, Kun Wang, Tzyy-Ping Jung, and Dong Ming.
\newblock Enhancing transfer performance across datasets for brain-computer interfaces using a combination of alignment strategies and adaptive batch normalization.
\newblock \emph{Journal of neural engineering}, 18\penalty0 (4):\penalty0 0460e5, 2021.

\end{thebibliography}
\bibliographystyle{iclr2024_conference}

\section{Appendix}

\subsection{Sensor failure performance across subjects}\label{appendix:sensor_failure_detailed}

\begin{figure}[h!]
    \centering
    \includegraphics[width=0.9\linewidth]{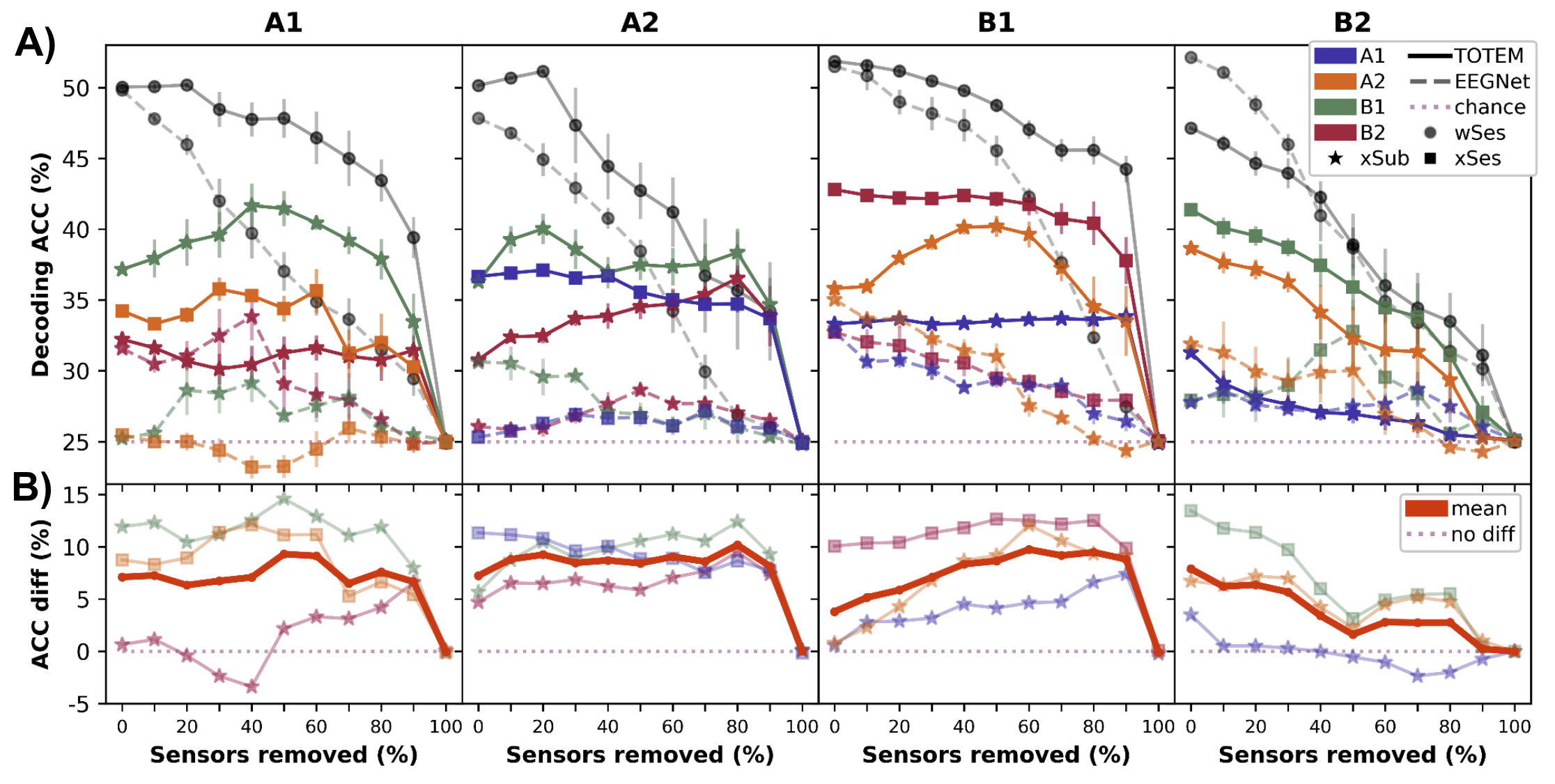}
    \caption{Classifier performance under sensor failure. A) Classifier accuracy across variable sensor failure \%s. Each column represents models trained on the dataset in the title. Error bars represent SEM of 5 random seeds which randomly select the failed sensors. wSes=Within Session, xSes=Cross Session, xSub=Cross Subject. B) TOTEM \& EEGNet accuracy difference with mean of differences in red. If a point lies above 0, TOTEM decodes better.}
    \label{fig:sensor_failure_detailed}
\end{figure}

% \vspace{5em} % Adjust the space here
\newpage

\subsection{Latent codebook generalizability across all cases}\label{appendix:codebook_generalizability}

\begin{figure}[h!]
    \centering
    \begin{subfigure}[t]{\textwidth}
        \includegraphics[width=\hsize]{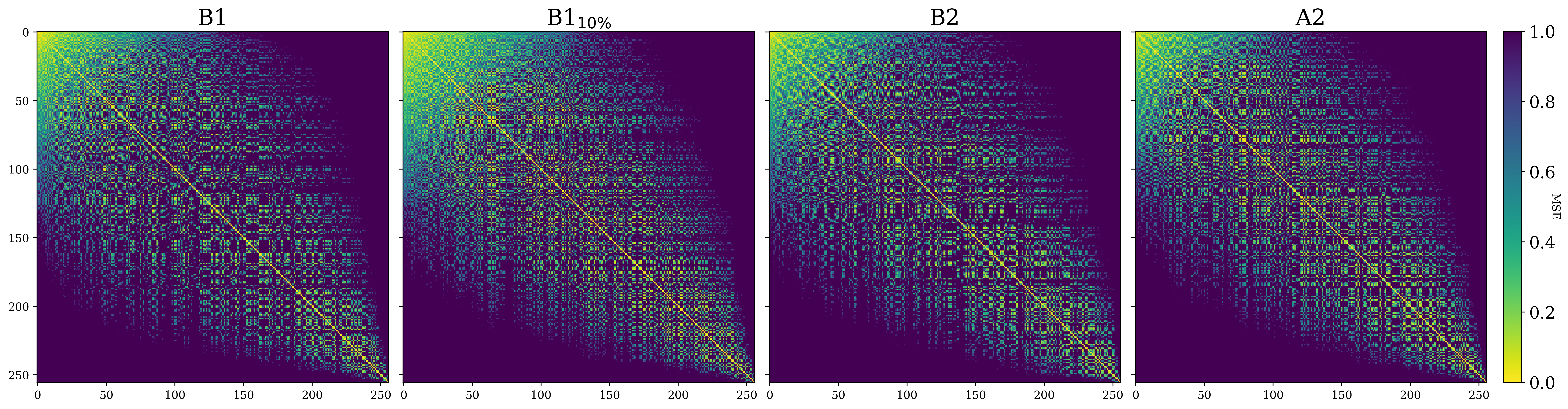}
        \caption{Within Session Codebook MSE Matrices}
        \label{fig:appendix_codebooks_mse_within}
    \end{subfigure}
    \vspace{2em} % Adjust the space here

    \begin{subfigure}[t]{\textwidth}
        \includegraphics[width=\hsize]{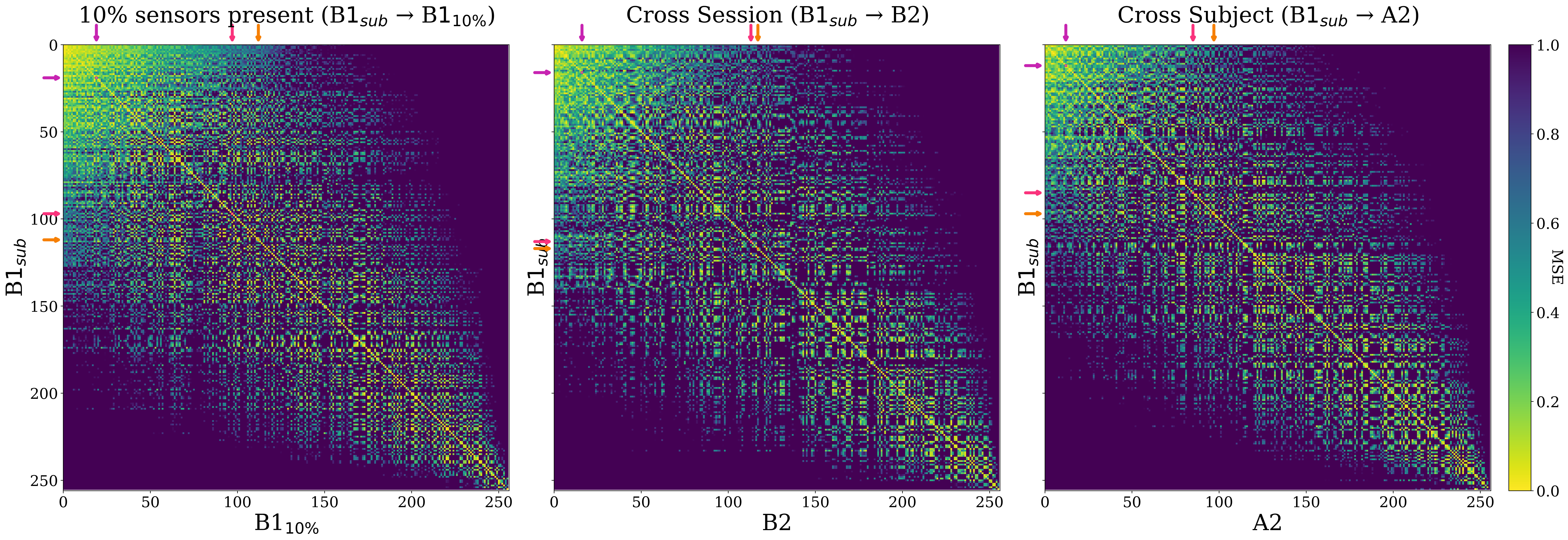}
        \caption{Matrices of MSE between matching codewords in the two labeled codebooks. Codewords are ordered roughly by within session mean MSE. Colored arrows along axis denote highlighted codewords in (c).}
        \label{fig:appendix_codebooks_mse_generalization}
    \end{subfigure}
    \vspace{2em} % Adjust the space here

    \begin{subfigure}[t]{\textwidth}
        \includegraphics[width=\hsize]{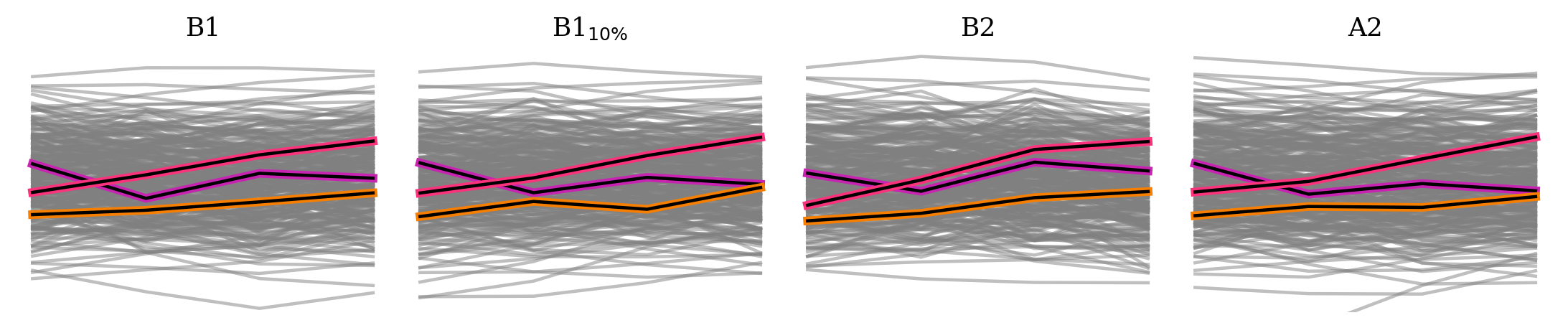}
        \caption{All codewords in each codebook. Highlighted are three examples of matched codewords between codebooks}
        \label{fig:appendix_codewords}
    \end{subfigure}
    \caption{Additional latent codebook visualizations}
    \label{fig:codebook_generalizability}
\end{figure}

\end{document}